\RequirePackage{fix-cm}
\documentclass[twocolumn]{svjour3}
\smartqed

\usepackage[utf8]{inputenc}
\usepackage[english]{babel}
\usepackage{graphicx}
\usepackage{epsfig}
\usepackage{booktabs}
\usepackage{multirow}
\usepackage{array}
\usepackage{amsmath}
\usepackage{amssymb}
\usepackage{amsfonts}
\usepackage{mathrsfs}
\usepackage{bm}
\usepackage{enumerate}
\usepackage[round,sort]{natbib}
\usepackage{setspace}
\usepackage[ruled,vlined]{algorithm2e}
\usepackage[dvipsnames,table]{xcolor}
\usepackage{colortbl}
\usepackage{arydshln}
\usepackage{url}
\usepackage{hyperref}
\usepackage{microtype}
\usepackage{marvosym}

\usepackage{amsmath,amsfonts,bm}

\def\eqref#1{equation~\ref{#1}}

\def\1{\bm{1}}

\DeclareMathAlphabet{\mathsfit}{\encodingdefault}{\sfdefault}{m}{sl}
\SetMathAlphabet{\mathsfit}{bold}{\encodingdefault}{\sfdefault}{bx}{n}

\journalname{}
\makeatletter
\def\makeheadbox{}
\makeatother

\begin{document}

\title{TivTok: Broadcasting Time-Invariant Tokens for Scalable Video Tokenization}
\titlerunning{TivTok}

\author{Weiliang~Chen \and
        Yuanhui~Huang \and
        Xuebo~Wang \and
        Yueqi~Duan
}
\authorrunning{W. Chen et al.}

\institute{Weiliang Chen\textsuperscript{1} \at
              \email{cwl24@mails.tsinghua.edu.cn} \\
           Yuanhui Huang\textsuperscript{2} \at
              \email{huangyh22@mails.tsinghua.edu.cn} \\
           Xuebo Wang\textsuperscript{3} \at
              \email{wangxuebo@kuaishou.com} \\
           Yueqi Duan\textsuperscript{1,\textrm{\Letter}} \at
              \email{duanyueqi@tsinghua.edu.cn} \\ \\
           \textsuperscript{1} Department of Electronic Engineering, Tsinghua University, Beijing, 100084, China\\
           \textsuperscript{2} Department of Automation, Tsinghua University, Beijing, 100084, China\\
           \textsuperscript{3} Kuaishou Technology, Beijing, China
}

\date{}

\maketitle

\begin{abstract}
Video tokenization is fundamental to scalable video generation, as the number of tokens directly determines the computational cost and the length of videos that can be modeled. Existing tokenizers mainly improve scalability by compressing videos into fewer tokens, but they often continue to represent persistent content, such as static backgrounds and consistent object appearances, repeatedly across frames and chunks. In this paper, we propose \textbf{TivTok} (\textit{Time-Invariant Tokenizer}), a reuse-aware video tokenizer that makes persistent information reusable across time. TivTok represents a clip with Time-Invariant (TIV) tokens that encode information shared across frames and Time-Variant (TV) tokens that encode frame-specific residuals. To obtain this factorization, we introduce Scope-Induced Factorization (SIF), which assigns different attention scopes to the two token groups: TIV tokens attend to the full clip, whereas each TV token only accesses its corresponding frame together with the TIV tokens. In the decoder, Invariant Broadcasting (IB) reuses the same TIV tokens across frames and chunks for parallel reconstruction and long-video tokenization. Experiments show that TivTok achieves an rFVD of 12.65 on the standard $16{\times}256{\times}256$ benchmark and improves compression efficiency by 2.91$\times$ for 128-frame videos compared with the evaluated baselines, while using only 1.1\% of the tokens required by downsample-based tokenizers in our evaluation.
\end{abstract}

\begin{figure*}[t]
    \centering
    \includegraphics[width=\textwidth]{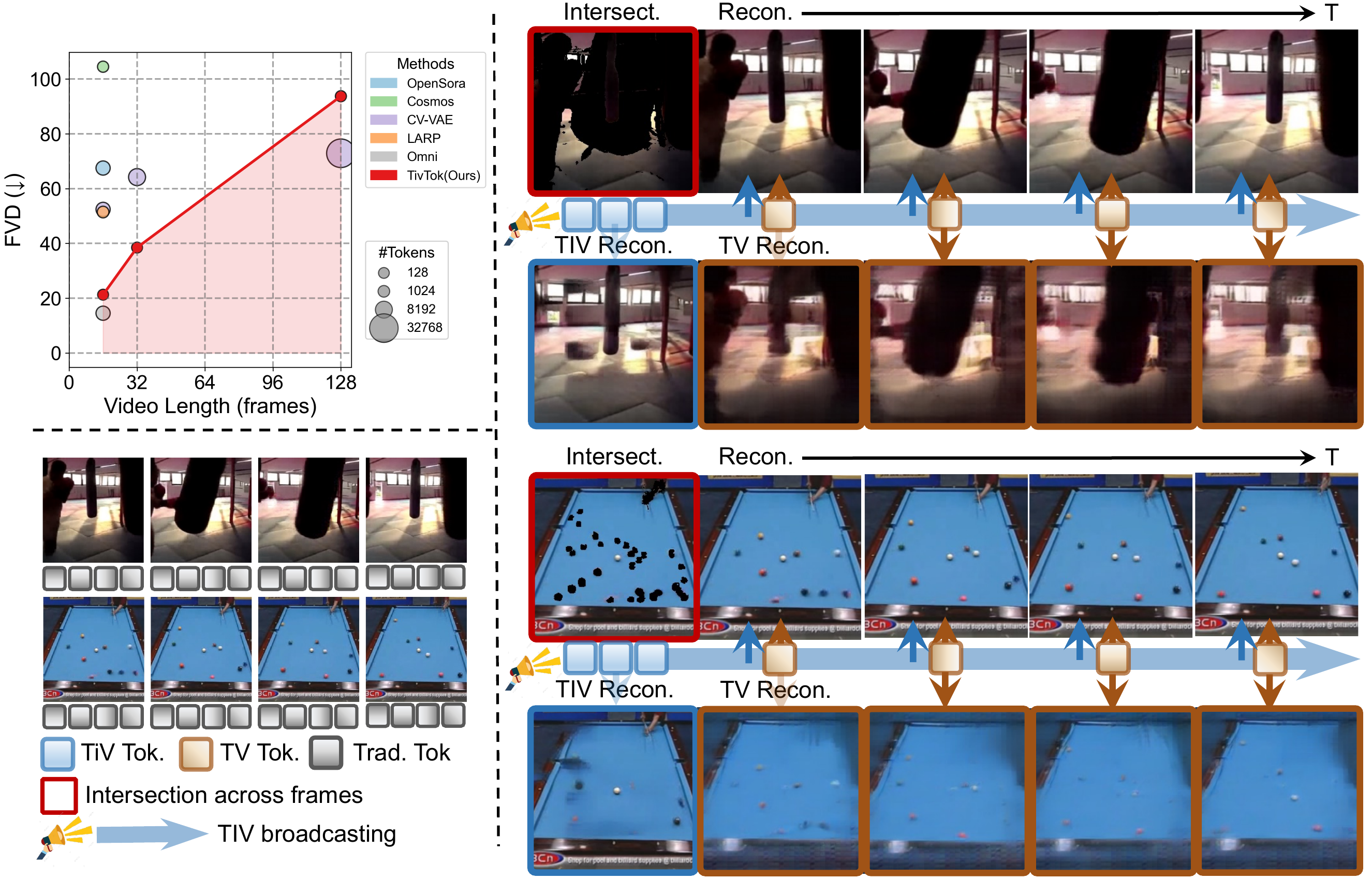}
    \caption{Overview of TivTok and its reuse-aware video tokenization. Top-left: reconstruction FVD is compared across video lengths, with marker size indicating the number of tokens; TivTok keeps a compact token budget while maintaining competitive reconstruction quality in long-video settings. Bottom-left: conventional tokenization treats persistent content and frame-specific variation uniformly when allocating representation capacity. Right: in contrast, the boxing and billiards examples illustrate how TivTok separates reusable Time-Invariant (TIV) tokens from frame-specific Time-Variant (TV) tokens. TIV tokens capture content shared over time, such as scene layout and object appearance, while TV tokens represent frame-specific changes such as object position and local motion. Broadcasting TIV tokens across frames and chunks allows persistent information to be reused rather than re-encoded at every frame.}
    \label{fig:teaser}
\end{figure*}

\section{Introduction}
Generative models have achieved remarkable success across diverse downstream applications, including visual content generation~\citep{blattmann2023stablevideodiffusion, rombach2022highlatentdiffusion, blattmann2023alignyourlatents, zhang2025scenariodiff, shu2026guidedvdm, liu2024makeyour3d, huang2025spectralar}, cinematic production~\citep{huang2024owl, chen2024dreamcinema, huang2025filmaster}, and industrial simulation~\citep{zheng2024occworld, ren2024xcube, ren2024scube, chen2025scenecompleter, agarwal2025cosmos}. A key factor behind this progress is compact visual tokenization: pixel-space visuals are highly redundant, and projecting them into lower-dimensional latent spaces significantly reduces computation and shifts focus to semantic structure, enabling sharper and higher-quality generations~\citep{rombach2022highlatentdiffusion, blattmann2023stablevideodiffusion, yang2021semantichierarchy}. However, video tokenization remains challenging, since videos introduce an additional temporal dimension, causing the amount of visual data and the number of tokens to grow with sequence length. At the same time, this temporal structure creates an opportunity for reuse: consecutive frames often share substantial content, suggesting that persistent information can be represented once and reused rather than re-encoded~\citep{xu2023conditionaltemporalvae}.

Consider what changes and what remains stable across consecutive frames. In many videos, scene layout, background appearance, and object identity remain largely consistent over time, while frame-specific factors such as object position, pose, and local motion vary from frame to frame. The billiards example in Fig.~\ref{fig:teaser} illustrates this structure: the table and ball appearances are shared across the sequence, whereas the ball positions and interactions change over time. Reusing the shared component becomes increasingly beneficial as video length grows, because information encoded once can be reused by an increasing number of frames. An effective tokenizer should therefore treat shared and changing content differently, encoding persistent information once while reserving per-frame capacity for temporal residuals.

Existing video tokenizers are largely compression-focused. Downsample-based methods extend image tokenizers with temporal modules or 3D convolutions~\citep{blattmann2023alignyourlatents, li2024hunyuan, agarwal2025cosmos, zhao2024cvvae, hacohen2024ltx}, while holistic tokenizers use transformers to compress video patches into compact latent tokens~\citep{yu2024imagetitok, wang2024larp, yan2024elastictok, li2025learningadaptok}. These designs reduce token counts, but they usually allocate representation capacity over the clip without explicitly separating reusable content from frame-specific variation. Another line introduces prescribed decompositions, such as reference frames with motion residuals~\citep{tan2024sweettok, tian2024reducio, yu2024efficientcmd, wang2025vidtwin} or frequency-based separation~\citep{liu2025hivae}. Such decompositions simplify the representation, but they are still primarily used for compression within a clip rather than for reusing persistent information across frames and chunks. This leaves room for a reuse-aware tokenizer that can discover persistent information and broadcast it over time.

Driven by this observation, we propose TivTok (\emph{Time-Invariant Tokenizer}), a reuse-aware video tokenizer that makes persistent information reusable across time. TivTok represents a clip with two types of tokens: Time-Invariant (TIV) tokens encode information shared across frames, while Time-Variant (TV) tokens encode frame-specific residuals, as illustrated in Fig.~\ref{fig:teaser}. TivTok realizes this factorization with two complementary mechanisms. In the encoder, Scope-Induced Factorization (SIF) assigns different attention scopes to the two token groups: TIV tokens attend to the full clip to aggregate shared information, whereas each TV token only accesses its corresponding frame together with the TIV tokens. This structural constraint encourages the model to place reusable information in TIV tokens and frame-specific residuals in TV tokens. In the decoder, Invariant Broadcasting (IB) reuses the same TIV tokens for every frame and combines them with the corresponding TV tokens for parallel reconstruction, reducing decoding complexity from $\mathcal{O}(T^2)$ to $\mathcal{O}(T)$ in video length. For long videos, TivTok reuses TIV tokens across chunks, allowing the shared representation to support a longer temporal range. With this design, TivTok achieves an rFVD of 12.65 in the standard $16{\times}256{\times}256$ setting and improves compression efficiency by $2.91{\times}$ for $128{\times}256{\times}256$ videos compared with baselines. Our main contributions are as follows:

\begin{itemize}
\item We formulate video tokenization from a reuse perspective, showing that persistent structure can be represented once and reused across frames and chunks. Based on this view, we propose TivTok, a reuse-aware tokenizer with Time-Invariant (TIV) and Time-Variant (TV) tokens.
\item We introduce Scope-Induced Factorization (SIF) and Invariant Broadcasting (IB). SIF uses asymmetric attention scopes to separate shared information from frame-specific residuals, while IB broadcasts TIV tokens for parallel frame reconstruction, reducing decoding complexity from $\mathcal{O}(T^2)$ to $\mathcal{O}(T)$.
\item We validate TivTok on video reconstruction and generation benchmarks, including long-video settings. TivTok achieves an rFVD of 12.65 on $16{\times}256{\times}256$ videos and improves compression efficiency by $2.91\times$ on $128{\times}256{\times}256$ videos, while using only 1.1\% of the tokens required by baselines.
\end{itemize}

\section{Related Work}

\subsection{From Image Tokenizer to Video Tokenizer} 
Following the progress of the encode-generate paradigm in image generation~\citep{rombach2022highlatentdiffusion}, researchers have developed video tokenizers by extending image tokenization methods to the temporal dimension. These approaches can be categorized into two main lines of work. \emph{Downsample-based video tokenizers.} Early work~\citep{blattmann2023alignyourlatents} adapts image tokenizers for video by encoding videos frame-by-frame. Subsequent methods~\citep{zhao2024cvvae, agarwal2025cosmos, chen2024odvae, tang2024vidtok} extend 2D convolutions to 3D for temporal downsampling, achieving higher compression ratios through various optimization techniques. CV-VAE~\citep{zhao2024cvvae} leverages image-pretrained 2D convolutions to regularize video tokenizers, improving training efficiency. VidTok~\citep{tang2024vidtok} incorporates FSQ to improve codebook utilization and compression efficiency. Cosmos~\citep{agarwal2025cosmos} employs 3D Haar wavelets to enhance model performance.

\emph{Holistic video tokenizers.} TiTok~\citep{yu2024imagetitok} pioneered transformer-based tokenization by compressing images into compact 1D learnable tokens via global receptive fields, inspiring subsequent works for images~\citep{huang2025spectralar, tian2024visualvar} and videos~\citep{wang2024larp, yan2024elastictok, li2025learningadaptok}. Directly applying such methods to videos remains challenging because the number of video patches grows with temporal length, increasing attention cost and limiting scalability in long-video settings. Both downsample-based and holistic tokenizers reduce token counts, but they typically allocate representation capacity over the clip without explicitly separating reusable content from frame-specific variation. TivTok follows a reuse-aware design by representing shared content with TIV tokens and frame-specific residuals with TV tokens.

\subsection{Decomposition-based Video Tokenizers}
Traditional video compression standards exploit temporal redundancy through decomposed encoding, as in H.264 and AV1~\citep{richardson2004hh264mpeg4,de2019av1}. P-frames encode residuals relative to previously decoded frames, reusing shared content across consecutive frames. Recent learned methods also build on temporal redundancy and decomposed representations~\citep{wu2024ivideogpt, jin2024videolavit, yu2024efficientcmd, wang2026breakingredundancy}. CMD~\citep{yu2024efficientcmd} encodes videos into a 2D content frame and low-dimensional motion latents. Reducio~\citep{tian2024reducio} uses an image-conditioned decoder with a reference image. SweetTok~\citep{tan2024sweettok} encodes the first frame and subsequent residual frames, while HiVAE~\citep{liu2025hivae} separates high- and low-frequency components.

Despite their similarity to P-frame coding, these methods primarily target compression within a single video. They simplify what each component represents, but do not explicitly reuse persistent structure across clips and chunks. In contrast, TivTok brings the reuse philosophy of H.264 into learned tokenization: Scope-Induced Factorization (SIF) discovers video-adaptive invariants that can be reused across frames and chunks.

\section{Method}
\subsection{Preliminary: Transformer-based Holistic Visual Tokenizer}
Pioneered by TiTok~\citep{yu2024imagetitok}, transformer-based holistic tokenizers have become a popular choice for visual tokenization. Their key idea is to distill a compact set of 1D global latents from all input patches by leveraging the transformer’s global receptive field.

Given a video $V \in \mathbb{R}^{3 \times T \times W \times H}$, the tokenizer first \emph{patchifies} $V$ with downsampling ratio $(f_T,f_W,f_H)$. This produces a grid of patch features $X$ with temporal-spatial size $(T/f_T)\times(W/f_W)\times(H/f_H)$ and channel dimension $d$. The flattened patches are then concatenated with learnable tokens $Z \in \mathbb{R}^{d \times N_z}$ to form $\tilde{Z} = [\operatorname{Flatten}(X);Z]$.

This combined sequence is then passed through a transformer encoder \(E(\cdot)\). Through self-attention, the latent tokens absorb global information from all patches across the video. After encoding, the latent tokens are quantized with \(Q(\cdot)\) to form a compact representation \(\hat{Z}\) that captures the essential content of the video in a discrete code space.

During decoding, learnable patch queries \(Q_p\) and the latent codes \(\hat{Z}\) are processed by a symmetric transformer decoder \(D(\cdot)\) to recover patch features
\(\hat{X} = D([Q_p;\hat{Z}])\),
which are then upsampled to the original resolution. 
This entire process can be summarized as  
\begin{equation}
\begin{aligned}
\hat{Z} &= \operatorname{Quant}\bigl(E(\tilde{Z})\bigr), \\
\hat{V} &= \operatorname{Unpatchify}\bigl(D([Q_p;\hat{Z}])\bigr).
\end{aligned}
\end{equation}
However, because the number of patches increases linearly with video length, the computational cost of self-attention grows quadratically; both encoding and decoding scale as $O(T^2)$. 

\subsection{Decoupling Time-Invariant and Time-Variant Tokens}
\begin{figure*}
    \centering
    \includegraphics[width=\textwidth]{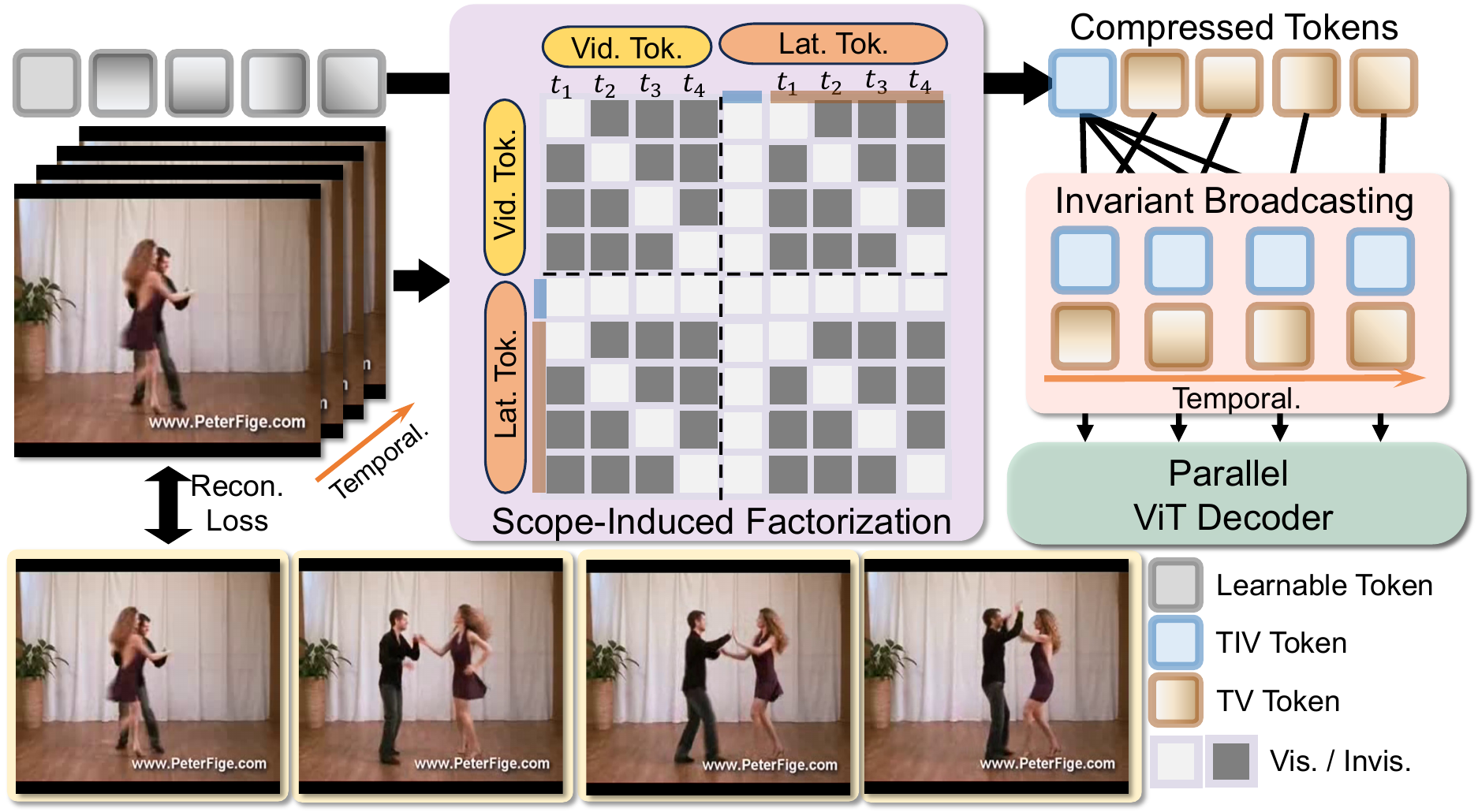}
    \caption{TivTok architecture overview. Given an input video, the encoder applies Scope-Induced Factorization (SIF) by assigning different attention scopes to the two token groups: TIV tokens attend to the full clip to aggregate shared information, while each TV token attends to its corresponding frame together with the TIV tokens to model frame-local variation. The compressed representation contains a shared set of TIV tokens and per-frame TV tokens. In the decoder, Invariant Broadcasting (IB) reuses the same TIV tokens at every time step and combines them with the corresponding TV tokens for parallel reconstruction.}
    \label{fig:pipeline}
\end{figure*}

We view the \textit{temporal invariant} of a video clip as the reusable component $C$ that is informative for multiple frames. This component is not limited to pixel-level static content; it can include semantically stable structure such as scene geometry, object identity, and consistent visual patterns that persist throughout the sequence. Encoding such information once and reusing it across frames can reduce repeated representation of scene-level content, leaving frame-specific residuals to be represented per frame.

This reuse can be motivated from an information-theoretic perspective. Suppose a video is represented by a shared component $C$ and per-frame residuals conditioned on $C$. Compared with encoding each frame separately, the amount of repeatedly encoded information can be written as
\begin{equation}
\label{equation:redundancy}
\begin{split}
H_{\text{indep}} - H_{\text{shared}}
&= \sum_{t=1}^{T} I(X_t; C) - H(C) \\
&\approx (T-1)H(C) > 0.
\end{split}
\end{equation}
where $H_{\text{indep}}$ denotes the sum of per-frame entropies, and $H_{\text{shared}}=H(C)+\sum_{t=1}^{T}H(X_t\mid C)$ corresponds to encoding the shared component once together with frame-wise residuals. The approximation holds when $C$ is recoverable from most frames. Under this condition, the potential saving grows with video length $T$, motivating a tokenizer that represents reusable content separately from frame-specific variation.

Building on this analysis, we factorize a video $V$ into two complementary token groups:\begin{itemize}
    \item \emph{Time-Invariant (TIV) tokens}, 
    $Z_{\text{TIV}} \in \mathbb{R}^{N_{\text{TIV}} \times D}$: 
    encode the shared semantic structure across all 
    frames, and can be reused to extend representations 
    to longer videos.
    \item \emph{Time-Variant (TV) tokens}, 
    $Z_{\text{TV}} \in \mathbb{R}^{T \times N_{\text{TV}} \times D}$: 
    preserve frame-specific residual details unique to 
    each time step.
\end{itemize}
A clip is represented as $[Z_{\text{TIV}}, Z_{\text{TV}}^{(1)}, \dots, Z_{\text{TV}}^{(T)}]$, while frame $t$ is reconstructed from $[Z_{\text{TIV}}, Z_{\text{TV}}^{(t)}]$. We enforce this factorization with SIF (Sec.~\ref{sec:sif}) and exploit it with IB (Sec.~\ref{sec:ib}).

\subsection{Scope-Induced Factorization}
\label{sec:sif}

Our encoder is guided by a single design principle: \textit{the information flow of a token should match its representational role}. A token intended to capture temporal invariants must see the entire sequence; a token intended to capture frame-specific residuals must be prevented from absorbing cross-frame information. We realize this through Scope-Induced Factorization (SIF), which enforces the TIV/TV decoupling through asymmetric attention scoping in the encoder.

Specifically, TIV tokens are granted global visibility: for a video $V = \{X_1, \dots, X_T\}$, each TIV token attends to all frame patches $\{X_t\}$ as well as all TV tokens. In contrast, each TV token at time step $t$ has only local visibility, restricted to its own frame patches $X_t$, the TIV tokens, and itself. We define two key-value scopes:
\begin{equation}
\begin{aligned}
\mathcal{G} &= [Z_{\text{TIV}}, Z_{\text{TV}}^{(1)}, \dots, Z_{\text{TV}}^{(T)}, X_1, \dots, X_T],\\
\mathcal{L}_t &= [Z_{\text{TIV}}, Z_{\text{TV}}^{(t)}, X_t]
\end{aligned}
\end{equation}
where $\mathcal{G}$ is the global scope visible to TIV queries and $\mathcal{L}_t$ is the frame-local scope visible to TV queries at time step $t$. We use $\operatorname{Attn}(A,B)$ to denote an attention update in which tokens in $A$ provide the queries and only tokens in $B$ are used as keys and values. The encoder updates are then written compactly as
\begin{equation}
\begin{aligned}
Z_{\text{TIV}}' &= \operatorname{Attn}(Z_{\text{TIV}}, \mathcal{G}), \\
Z_{\text{TV}}^{(t)\prime} &= \operatorname{Attn}(Z_{\text{TV}}^{(t)}, \mathcal{L}_t).
\end{aligned}
\end{equation}
This asymmetric scoping encourages TIV tokens to aggregate shared information across frames while keeping TV tokens focused on frame-local residuals, so the TIV/TV factorization is induced by the architecture rather than imposed through explicit supervision. Using causal masking for TV tokens may appear suitable for autoregressive generation, but it would allow TV tokens to absorb cross-frame information that overlaps with the TIV tokens and would raise encoding cost to quadratic in $T$. Restricting TV tokens to single-frame visibility keeps the token roles separated and reduces encoding complexity from $\mathcal{O}\!\left(T^2 \cdot (N_{\text{TIV}} + N_{\text{TV}})\right)$ to $\mathcal{O}\!\left(T^2 \cdot N_{\text{TIV}} + T \cdot N_{\text{TV}}\right)$.

\subsection{Invariant Broadcasting}
\label{sec:ib}

Having extracted TIV tokens that capture shared semantics across all frames, we design the decoder to exploit this structure directly through Invariant Broadcasting (IB). Rather than decoding frames sequentially, the same TIV tokens are broadcast to every time step and combined with the corresponding TV tokens, so that each frame is decoded as
\begin{equation}
\hat{X}_t = D\bigl([Z_{\text{TIV}}, Z_{\text{TV}}^{(t)}]\bigr),
\quad t=1,\dots,T.
\end{equation}
where $D(\cdot)$ denotes the transformer decoder. Since each frame's decoding depends only on the shared TIV tokens and its own TV tokens, all frames can be reconstructed in parallel. This design reduces decoding complexity from $\mathcal{O}(T^2)$ to $\mathcal{O}(T)$ in video length and supports long-video tokenization through cross-chunk TIV reuse, as described in the following section.

\subsection{Cross-Chunk TIV Reuse for Long Video 
Tokenization}
\label{sec:long_video}

The reuse motivation in Eq.~\ref{equation:redundancy} also applies to long videos. For a $K$-chunk video, some persistent content can remain shared across chunks, so representing it independently in each chunk can introduce repeated tokens. TivTok reduces this repetition by reusing a single set of TIV tokens across all chunks. Specifically, for a long video $\{X_{1:TK}\}$ composed of $K$ chunks of length $T$, we first encode all $K$ chunks in parallel and merge their TIV tokens by averaging:
\begin{equation}
\bar{Z}_{\text{TIV}} = \frac{1}{K}\sum_{i=1}^{K} Z_{\text{TIV}}^{(i)}.
\end{equation}
The merged TIV tokens $\bar{Z}_{\text{TIV}}$ capture the global 
shared semantics of the entire video, and the full 
representation is reorganized as:
\begin{equation}
\mathcal{Z} = [\bar{Z}_{\text{TIV}},\{Z_{\text{TV}}^{(i,t)}\}_{i=1,\ldots,K;\,t=1,\ldots,T}].
\end{equation}
where $Z_{\text{TV}}^{(i,t)}$ denotes the TV tokens of the $t$-th frame in chunk $i$. During decoding, $\bar{Z}_{\text{TIV}}$ is broadcast to every frame following the IB mechanism (Sec.~\ref{sec:ib}), enabling parallel reconstruction across all chunks. The full training procedure is detailed in Algorithm~\ref{alg:propagation}. This design yields three concrete benefits: it reduces total token count by eliminating redundant TIV tokens across chunks; it cuts computational complexity from $\mathcal{O}(K^2)$ to $\mathcal{O}(K)$ in the number of chunks; and it eases optimization by shortening the effective token sequence length during training.

\begin{algorithm}[t]
\caption{Cross-Chunk TIV Reuse Training for Long 
Video Tokenization}
\label{alg:propagation}
\textbf{Input:} Long video $\{X_{1:TK}\}$ with $K$ 
chunks of length $T$\;
\textbf{Output:} Reconstructed video 
$\hat{X}_{1:TK}$\;

\textbf{1. Parallel Encoding:}\\
\For{$i = 1,\dots,K$}{
    Encode chunk $X^{(i)}_{1:T} \rightarrow 
    Z_{\text{TIV}}^{(i)},\ \{Z_{\text{TV}}^{(i,t)}\}_{t=1}^{T}$
}

\textbf{2. TIV Token Merging:}\\
$\bar{Z}_{\text{TIV}} = \dfrac{1}{K}\sum_{i=1}^{K} 
Z_{\text{TIV}}^{(i)}$\\

\textbf{3. Token Reorganization:}\\
$\mathcal{Z} = [\bar{Z}_{\text{TIV}},\{Z_{\text{TV}}^{(i,t)}\}_{i,t}]$\\

\textbf{4. Invariant Broadcasting (IB):}\\
\For{each frame $(i,t)$ \textbf{in parallel}}{
    $\hat{X}^{(i,t)} = D\bigl([\bar{Z}_{\text{TIV}},\ 
    Z_{\text{TV}}^{(i,t)}]\bigr)$
}

\textbf{5. Update:}\\
Compute $\mathcal{L}(X, \hat{X})$, update 
parameters\;

\textbf{Complexity:} $\mathcal{O}(K^2) \rightarrow 
\mathcal{O}(K)$\;
\end{algorithm}

\section{Experiments}

\subsection{Implementation Details}
Our tokenizer is built upon a ViT-based SoftVQ-VAE~\citep{chen2025softvq}. Unless otherwise specified, the encoder and decoder use 12 layers with hidden dimension 768, patch size $4\times8\times8$, and 3D RoPE positional embeddings. All tokenizers are trained on a mixture of UCF-101~\citep{soomro2012ucf101} and K600~\citep{carreira2018shortk600} for 100K iterations at $256\times256$ resolution. For long-video tokenization, we conduct an additional 50K iterations for cross-chunk TIV reuse. We use AdamW with weight decay $10^{-4}$, momentum $(\beta_1,\beta_2)=(0.9,0.95)$, global batch size 64, base learning rate $10^{-4}$, 5K warmup steps, and cosine learning-rate decay. Standard horizontal flipping and center cropping are used for data augmentation.

Our model $\phi$ is optimized using a composite loss function that combines reconstruction quality with perceptual and adversarial objectives:
\begin{equation}
\begin{split}
L &= L_{\text{recon}} + \lambda_1 L_{\text{percept}}
   + \lambda_2 \lambda_{\nabla} L_{\text{adv}},\\
\lambda_{\nabla} &=
\frac{\nabla_\phi (L_{\text{recon}} + \lambda_1 L_{\text{percept}})}
{\nabla_\phi L_{\text{adv}}}.
\end{split}
\end{equation}
This objective incorporates L1 reconstruction loss $L_{\text{recon}}$, perceptual loss $L_{\text{percept}}$~\citep{johnson2016perceptualloss, larsen2016autoencodingperceptualloss}, and adversarial loss $L_{\text{adv}}$~\citep{goodfellow2020generativeadversualloss}. We set $\lambda_1 = 1$ and $\lambda_2 = 0.2$, use a DINOv2-S discriminator starting from 30K iterations, and set LeCAM regularization to 0.001.

For video generation, we adapt the LightningDiT architecture~\citep{yao2025lightningdit} for class-conditional generation on UCF-101. The generation model uses hidden dimension 1152, 28 layers, 16 attention heads, patch size 1, and absolute positional embeddings. It is trained for 100K iterations with AdamW, global batch size 512, learning rate $10^{-4}$, constant learning-rate schedule, and center cropping. During inference, we use the Euler sampler with 50 diffusion steps, CFG interval start 0.1, and timestamp shift 2. For reconstruction, we report PSNR, SSIM~\citep{wang2004ssim}, LPIPS~\citep{zhang2018unreasonablelpips}, and reconstruction FVD (rFVD)~\citep{unterthiner2018towardsrfvd}. For generation, we report generation FVD (gFVD).

\subsection{Video Reconstruction Comparison}

\begin{table*}[t]
    \caption{Comparison of Video Reconstruction on UCF-101. 
    We compare different categories of video tokenizers with similar compression ratios. We additionally report the token-to-pixel ratio (T/P (\%)) for intuitive comparison, which is crucial for generation model efficiency. Bold values indicate best performance; \underline{underlined} values show second-best results.}
    \label{tab:main_results}
    \centering
    \setlength{\tabcolsep}{3.5pt}
    \begin{tabular*}{\textwidth}{@{\extracolsep{\fill}}lccccccc}
        \toprule
        \textbf{Method} & \textbf{\#Tokens} & \textbf{\#Dim.} & \textbf{T/P (\%)}$\downarrow$& \textbf{PSNR}$\uparrow$ & \textbf{SSIM}$\uparrow$ & \textbf{LPIPS}$\downarrow$  & \textbf{rFVD}$\downarrow$ \\
        \midrule
        \multicolumn{8}{l}{\textit{Downsample-based video tokenizer}} \\
        \hdashline 
        \addlinespace
        SDXL-VAE~\citep{podell2023sdxl} & 16384 & 4 &1.563 &- &- & - &23.68 \\
        OpenSora~\citep{zheng2024opensora} &4096 & 16 &0.391 & - &- &- &67.52 \\
        Cosmos-M~\citep{agarwal2025cosmos} &2048 & 16 &0.195 & \textbf{31.70} &\underline{0.9177} &0.0575 &\underline{13.67} \\
        Cosmos-S~\citep{agarwal2025cosmos} &512 & 16 &\underline{0.049}& 28.26 &0.8577 &0.1046 &104.51 \\
        CV-VAE~\citep{zhao2024cvvae} &4096 & 4 &0.391 & 29.47 &0.8849 &0.0685 &52.43 \\
        
        \midrule
        \multicolumn{8}{l}{\textit{Holistic video tokenizer}($^{*}$:Video resolution 16$\times$128$\times$128)} \\
        \hdashline 
        \addlinespace
        LARP~\citep{wang2024larp}$^{*}$ & 1024 &16 &0.391 & 28.65 &0.9003 &\textbf{0.0425} &23.93 \\
        LARP~\citep{wang2024larp} & 1024 &16 &0.098 & 25.53 &0.8262 &0.0973 &51.45 \\
        ElasticTok~\citep{yan2024elastictok} &1024 & 16 &0.391 &- &- &- &390 \\
        AdapTok~\citep{li2025learningadaptok} & 2048 & 16 &0.781 & 26.38 &0.8539 &0.0599 &27.97 \\
        
        \midrule
        \multicolumn{8}{l}{\textit{Decomposition-based video tokenizer}} \\
        \hdashline 
        \addlinespace
        Omni~\citep{wang2024omnitokenizer}  & 4096 & 8 &0.391 & 29.34 & \textbf{0.9250} & \underline{0.0487} & 14.53 \\
        Omni-DV~\citep{wang2024omnitokenizer} & 4096 & 8 &0.391 & 28.06 & 0.9095 & 0.0637 & 27.12 \\
        VidTwin~\citep{wang2025vidtwin} & 1008 & 4/8 &0.126 & 28.14 & 0.8044 & 0.2414 & 388.86\\
        \midrule
        TivTok-T128 & 128 & 128 &\textbf{0.012} &30.13 &0.9010 &0.0614 &21.29 \\
        TivTok-T512  & 512 & 32 &\underline{0.049}& \underline{30.26} & 0.8982 &0.0533 &\textbf{12.65} \\
        TivTok-T1024 & 1024 & 16 &0.098 & 29.54 & 0.8897 &0.0607 &17.97 \\
        \bottomrule
    \end{tabular*}
\end{table*}

We evaluate video reconstruction quality on UCF-101~\citep{soomro2012ucf101} at $256{\times}256$ resolution with 16-frame sequences, comparing against representative baselines from three categories: downsample-based, holistic, and decomposition-based tokenizers, all configured at similar compression ratios for meaningful comparison.

As shown in Table~\ref{tab:main_results}, TivTok achieves competitive reconstruction quality with a much smaller token budget, and TivTok-T512 obtains the lowest rFVD among the evaluated $256{\times}256$ settings. Notably, TivTok-T512 achieves the best reconstruction quality among our model variants, suggesting a favorable balance between token count and dimensionality: too few tokens limit spatial resolution while overly low-dimensional tokens restrict representational capacity. The small differences across TivTok-T128/T512/T1024 indicate that the framework is stable across this trade-off, offering flexible operating points for different efficiency requirements.

Compared with decomposition-based methods, TivTok achieves stronger compression on both UCF-101 (Table~\ref{tab:main_results}) and WebVid (Table~\ref{tab:scalability}). We attribute this to the emergent nature of our TIV/TV factorization: rather than restricting the invariant component to a prescribed definition such as ``motion'' or ``high-frequency'', Scope-Induced Factorization discovers video-adaptive invariants that generalize across diverse content, leading to more efficient and accurate redundancy elimination.

\subsection{Long Video Tokenization}

\begin{table*}[t]
    \caption{Comparison of long video tokenization. We retrain baseline methods under their evaluated tokenization settings and compare against CoordTok~\citep{jang2025efficientcoordtok}. We report reconstruction metrics together with inference latency for computational efficiency assessment.}
    \label{tab:longvideocomparison}
    \centering
    \setlength{\tabcolsep}{3.5pt}
    \begin{tabular*}{\textwidth}{@{\extracolsep{\fill}}lccccccc}
        \toprule
        \textbf{Method} & \textbf{\#Tokens} & \textbf{\#Dim.} &\textbf{Latency (s)}$\downarrow$& \textbf{PSNR}$\uparrow$ & \textbf{SSIM}$\uparrow$ & \textbf{LPIPS}$\downarrow$  & \textbf{rFVD}$\downarrow$ \\
        \midrule
        \multicolumn{8}{l}{\textit{Video resolution 32$\times$256$\times$256}} \\
        \hdashline 
        \addlinespace
        CV-VAE~\citep{zhao2024cvvae} & 8192 & 4 & 1.78& 29.12 & 0.8809 & \underline{0.0692} & 64.21\\
        LARP~\citep{wang2024larp} & 2048 &16 &\underline{1.75} & 23.15 & 0.7479 & 0.1757 & 226.79 \\
        TivTok-T128 & 160 & 128 &\textbf{0.20} &29.05 &0.8831 &0.0719 & \underline{38.49} \\
        TivTok-T512 & 640 & 32 & - & \textbf{30.25} & \textbf{0.8948} & \textbf{0.0591} & 
        \textbf{23.26}\\
        TivTok-T1024 & 1280 & 16 & - & \underline{29.13} & \underline{0.8857} & 0.0711 & 61.46 \\
        \midrule
        \multicolumn{8}{l}{\textit{Video resolution 128$\times$256$\times$256
        ($^{*}$:Video resolution 128$\times$128$\times$128)}} \\
        \hdashline
        \addlinespace
        CV-VAE~\citep{zhao2024cvvae} & 32768 & 4 &\underline{7.12} & \textbf{29.00} & \textbf{0.8831} & \textbf{0.0729} & \textbf{72.91}\\
        LARP~\citep{wang2024larp} & 8192 &16 &22.78 & 14.85 & 0.2924 & 0.6251 &3223.55\\
        CoordTok~\citep{jang2025efficientcoordtok}$^{*}$ & 1280 & 8 &- & 27.25 & 0.7503 & 0.2346 & 1108.76 \\
        TivTok-T128 & 352 & 128 &\textbf{0.71} & 26.23 &\underline{0.8210} &\underline{0.1057} &\underline{92.09} \\
        \bottomrule
    \end{tabular*}
\end{table*}

\begin{figure*}
    \centering
    \includegraphics[width=\textwidth]{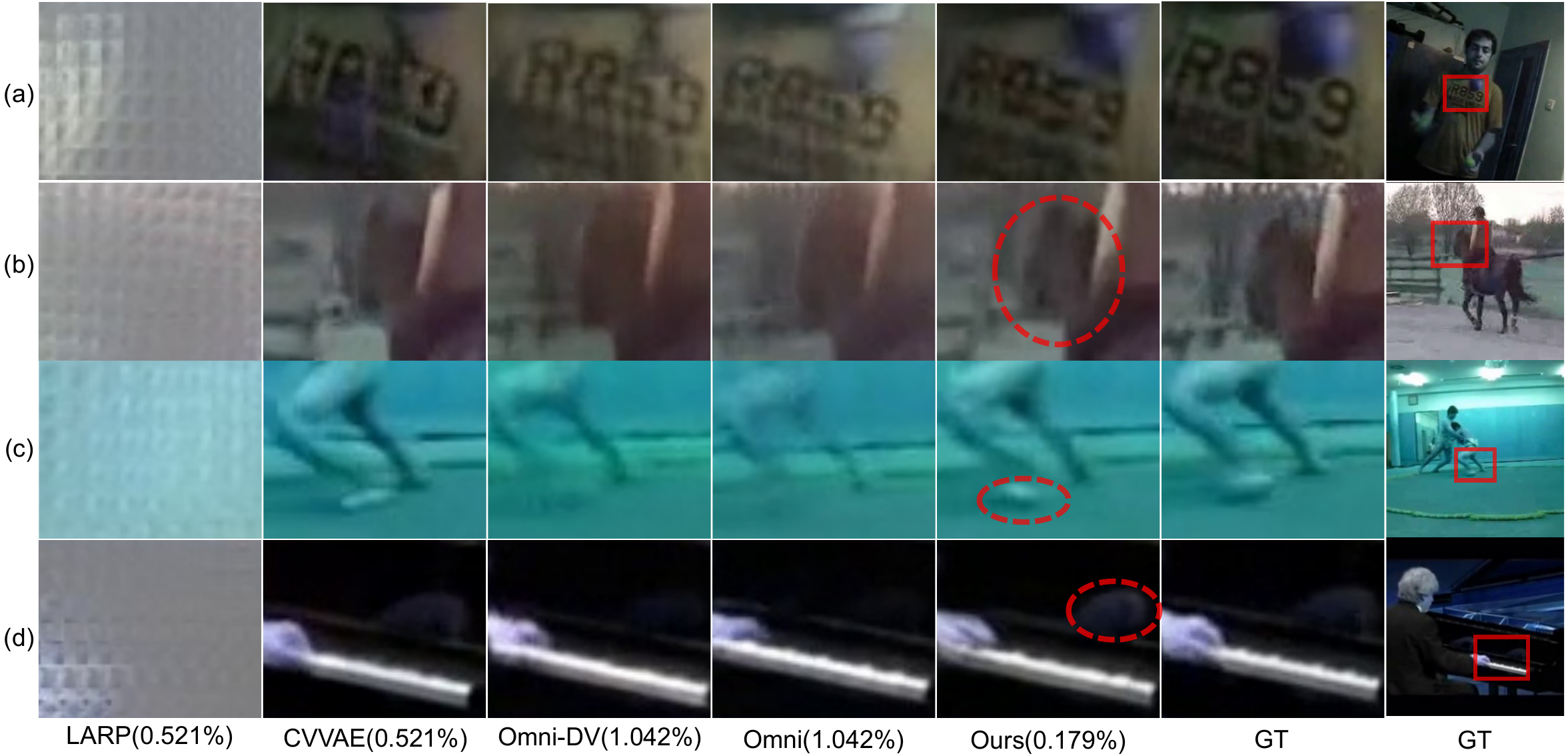}
    \caption{Long Video Reconstruction Comparison on UCF-101. Compression ratios are shown in parentheses (lower is better). Our method operates at a significantly lower compression ratio than baselines, yet demonstrates superior detail preservation, as highlighted by the red circles in the magnified regions.}
    \label{fig:recon}
\end{figure*}

\begin{figure*}[t]
    \centering
    \includegraphics[width=\textwidth]{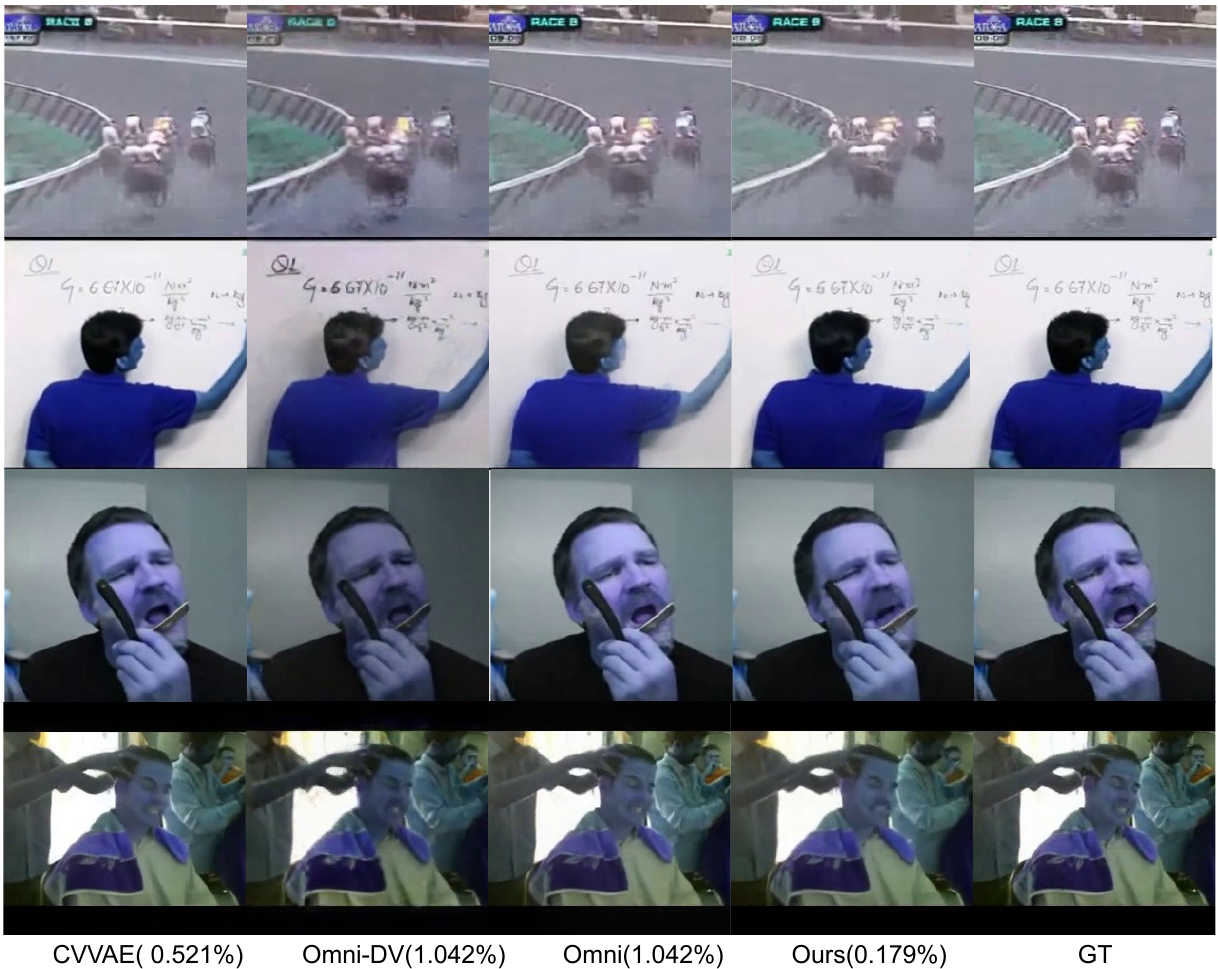}
    \caption{Full-frame reconstruction comparison on UCF-101. Compression ratios are shown in parentheses (lower is better). This figure complements the magnified details in Fig.~\ref{fig:recon} by showing complete reconstructed frames. TivTok preserves coherent global appearance while using substantially fewer tokens than the baselines.}
    \label{fig:morevis}
\end{figure*}

To evaluate temporal invariant reuse in longer sequences, we explore long video tokenization. The experimental results in Table~\ref{tab:longvideocomparison} reveal distinct behavioral patterns as temporal length $T$ increases. Downsample-based video tokenizers such as CV-VAE~\citep{zhao2024cvvae} maintain relatively stable reconstruction quality, but their token counts grow with video length. Holistic video tokenizers such as LARP~\citep{wang2024larp} use fewer tokens than downsample-based tokenizers, but show degraded reconstruction quality and higher latency at longer lengths. In contrast, TivTok reuses TIV tokens across chunks and achieves $2.91{\times}$ higher compression efficiency for 128-frame videos compared with the evaluated baselines. In our evaluation, TivTok uses 1.1\% of the tokens required by downsample-based methods, indicating its potential for improving generation efficiency under long-video settings. 

We further provide qualitative comparisons at a resolution of 128\,$\times$\,256\,$\times$\,256 from two complementary views. Figure~\ref{fig:morevis} shows full-frame reconstructions, where TivTok preserves coherent global appearance despite using a substantially lower compression rate than the baselines. Figure~\ref{fig:recon} then zooms into local details, showing the retained numerical text and ball in (a), the horse head in (b), the fine details around the foot in (c), and the subtle hand reflection on the piano surface in (d). Together, the full-frame and magnified visualizations show that TivTok preserves global coherence and selected local details under a smaller token budget, supporting the benefit of reusing temporal invariants in long-video reconstruction.

\subsection{Video Generation Comparison}

\begin{table*}[t]
\centering
\caption{Comprehensive Comparison of Video Generation.  
The comparison includes inference speed, GPU memory usage, computational cost (TFLOPs), and generation quality (FVD). Results of MeBT~\citep{yoo2023towardsMeBT}, PVDM~\citep{yu2023videopvdm}, HVDM~\citep{kim2024hybridhvdm}, CoordTok~\citep{jang2025efficientcoordtok}+SiT-L/2~\citep{ma2024sit} are taken from MALT~\citep{yu2025maltdiffusion}. ($^{*}$: Video resolution 128$\times$128$\times$128).}
\label{tab:Generation}
\setlength{\tabcolsep}{2pt}
\begin{tabular*}{\textwidth}{@{\extracolsep{\fill}}lcccccc}
    \toprule
    \textbf{Method}  & \textbf{Len.} & \textbf{\#Tokens} & \textbf{Time/Step (s)$\downarrow$}  & \textbf{Mem. (GB)$\downarrow$} & \textbf{TFLOPs$\downarrow$} & \textbf{FVD$\downarrow$} \\
    \midrule
    Cosmos-S & 16 & 512 & 0.047 & 2.62 & 0.49 & 191 \\
    Omni     & 16 & 4096 & 0.437 & 4.69 & 5.82 & 191 \\
    LARP     & 16 & 1024 & 0.083 & 2.73 & 1.05 & 107 \\   
    CV-VAE    & 16 & 4096 & 0.437 & 4.69 & 5.82 & 262\\   
    
    TivTok-T1024 & 16 & 1024 & 0.083 & 2.73 & 1.05& \textbf{99} \\
    TivTok-T512 & 16 & 512 & 0.047 & 2.62 & 0.49& 101 \\
    TivTok-T128 & 16 & \textbf{128} & \textbf{0.021} & \textbf{2.58} & \textbf{0.12} & 149 \\
    \midrule
    CV-VAE    & 32 & 8192 & 1.261 & 10.82 & 15.97 & 370 \\   
    TivTok-T128 & 32 & \textbf{160} & \textbf{0.021} & \textbf{2.58} & \textbf{0.15} & \textbf{300} \\
    \midrule
    MeBT$^{*}$ & 128 & 8192 & 6.53 & 13.3 & - &968 \\
    PVDM$^{*}$ & 128 & 16384 & 0.26 & 4.33 & -& 505 \\
    HVDM$^{*}$ & 128 & 32768 &1.514 &12.1 &- & 550 \\
    CoordTok+SiT-L/2$^{*}$ & 128 &1280 &- &- &- &369 \\
    MALT$^{*}$ & 128 & 4096 & - &- &- &220\\
    TivTok-T128$^{*}$ & 128 & \textbf{352} & \textbf{0.031} & \textbf{2.60} & \textbf{0.33} & \textbf{208} \\
    TivTok-T128 & 128 & \textbf{352} & \textbf{0.031} & \textbf{2.60} & \textbf{0.33} & 316 \\
    \bottomrule
\end{tabular*}
\end{table*}

Table~\ref{tab:Generation} reports the generation performance for class-conditional video synthesis on UCF-101~\citep{soomro2012ucf101} under different video lengths. Unless otherwise specified, all videos are generated at a resolution of $256\times256$. Under the conventional 16-frame setting, TivTok-T1024/T512 achieve lower FVD than the evaluated baselines while maintaining competitive computational cost. By further reducing the number of tokens, TivTok-T128 improves generation efficiency yet still retains competitive FVD performance, demonstrating a favorable trade-off between efficiency and visual quality. 
These efficiency advantages become more pronounced as the video length increases. While existing methods require rapidly growing token counts, computational cost, and memory consumption when scaling to longer sequences, TivTok maintains a compact token representation and stable resource usage. As a result, it reduces time, memory, and TFLOPs in the evaluated long-video settings while keeping FVD competitive.

\subsection{Analysis of Time-Invariant Tokens}
\begin{figure*}
    \centering
    \includegraphics[width=\textwidth]{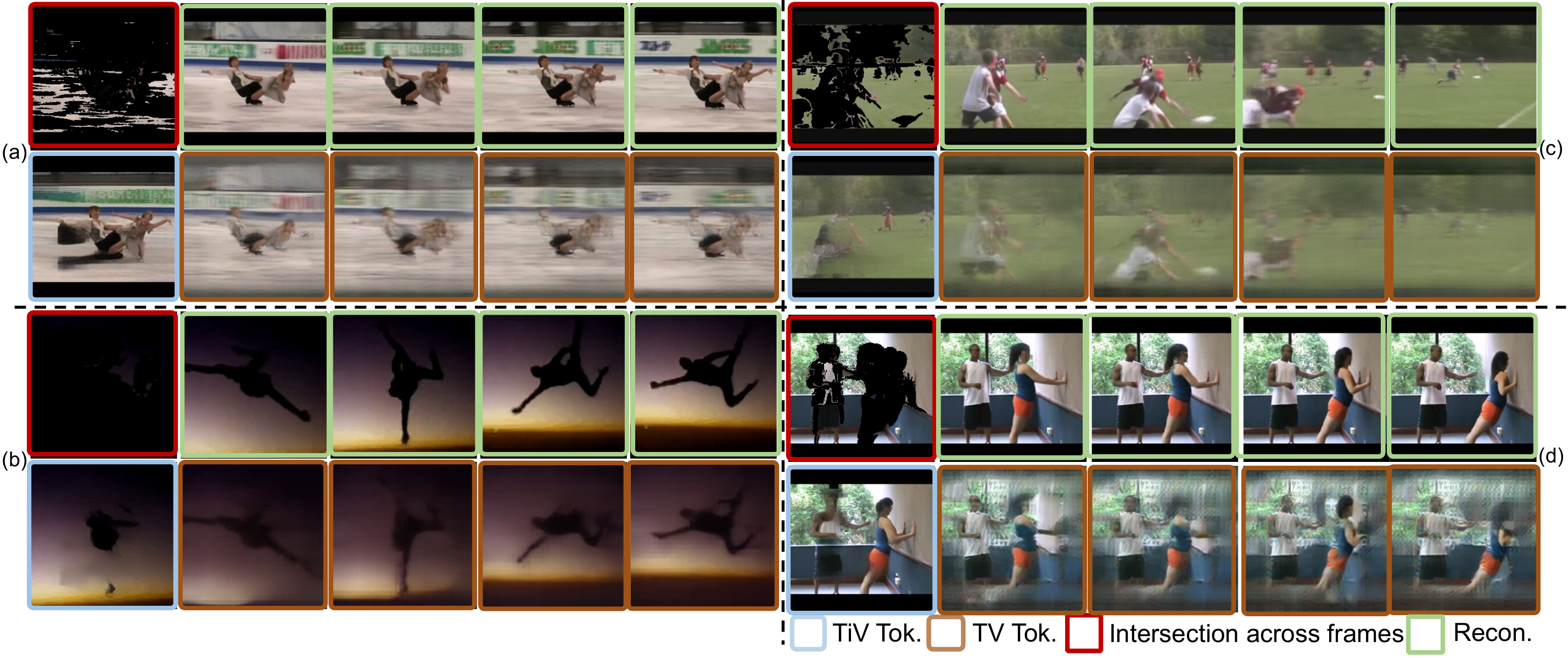}
    \caption{TIV Token and TV Token Visualization and Analysis. The intersection images (red boxes) display pixel-level persistence across frames, where we retain regions with minimal pixel variation. Results demonstrate that our TIV tokens capture temporal invariants including semantic information and scene geometry rather than merely pixel-level persistence.}
    \label{fig:token}
\end{figure*}

A key property of TIV tokens is that they capture \textit{semantic invariants} rather than pixel-level persistence. To illustrate this, we compare TIV token reconstructions against a simple pixel-intersection baseline, retaining only regions with minimal pixel variation across frames (red boxes in Figure~\ref{fig:token}). If TIV tokens merely encoded static pixels, their reconstructions would align closely with these intersections. Our results show otherwise.

In Figure~\ref{fig:token}(a), the background advertising boards change position across frames while the two skaters remain consistent, yet TIV tokens faithfully capture the skaters' detailed appearance and clothing textures rather than the pixel-stable background. In the pool-table example in Figure~\ref{fig:teaser}, stationary balls are captured as invariant while moving balls are delegated to TV tokens. These examples show that Scope-Induced Factorization discovers what is semantically stable rather than pixel-static.

Importantly, this behavior is not a foreground-background or motion-static split. A moving object can still be assigned to the invariant branch when its identity and appearance remain stable, while TV tokens only need to describe pose, location, and other transient changes. This flexibility distinguishes SIF from hand-crafted decompositions based on reference frames, motion masks, or frequency bands, and makes the learned invariant more suitable for reuse.

This emergent factorization has a direct practical consequence: TIV tokens provide reusable reconstruction priors, while TV tokens focus on temporal residuals. Such decoupling explains why TIV tokens can be broadcast across chunks and why the compression gain becomes more pronounced in long videos.

\subsection{Invariance Capacity and Token Allocation}
\begin{table*}[t]
\centering
\caption{Quantitative effect of the number of TIV tokens. Experiments are conducted at $128\times256\times256$ resolution; the table reports exact reconstruction and compression metrics.}
\label{tab:multipletiv}
\setlength{\tabcolsep}{3.5pt}
\begin{tabular*}{\textwidth}{@{\extracolsep{\fill}}lcccccccc}
\toprule
Method & Num TIV & Tokens & Dim & Comp. Rate (\%)$\downarrow$ & PSNR $\uparrow$ & SSIM $\uparrow$ & LPIPS $\downarrow$ & rFVD $\downarrow$ \\
\midrule
CV-VAE & - & 32768 & 4 & 0.521 & 29.00 & 0.8831 & 0.0729 & 72.91 \\
TivTok & 8 & 1024 & 128 & 0.521 & 30.07 & 0.9003 & 0.0618 & 28.96 \\
TivTok & 4 & 640 & 128 & 0.326 & 28.97 & 0.8825 & 0.0739 & 39.84 \\
TivTok & 2 & 448 & 128 & 0.228 & 27.20 & 0.8453 & 0.0951 & 81.18 \\
TivTok & 1 & 352 & 128 & 0.179 & 26.23 & 0.8210 & 0.1057 & 92.09 \\
\bottomrule
\end{tabular*}
\end{table*}

\begin{figure}[t]
    \centering
    \includegraphics[width=\linewidth]{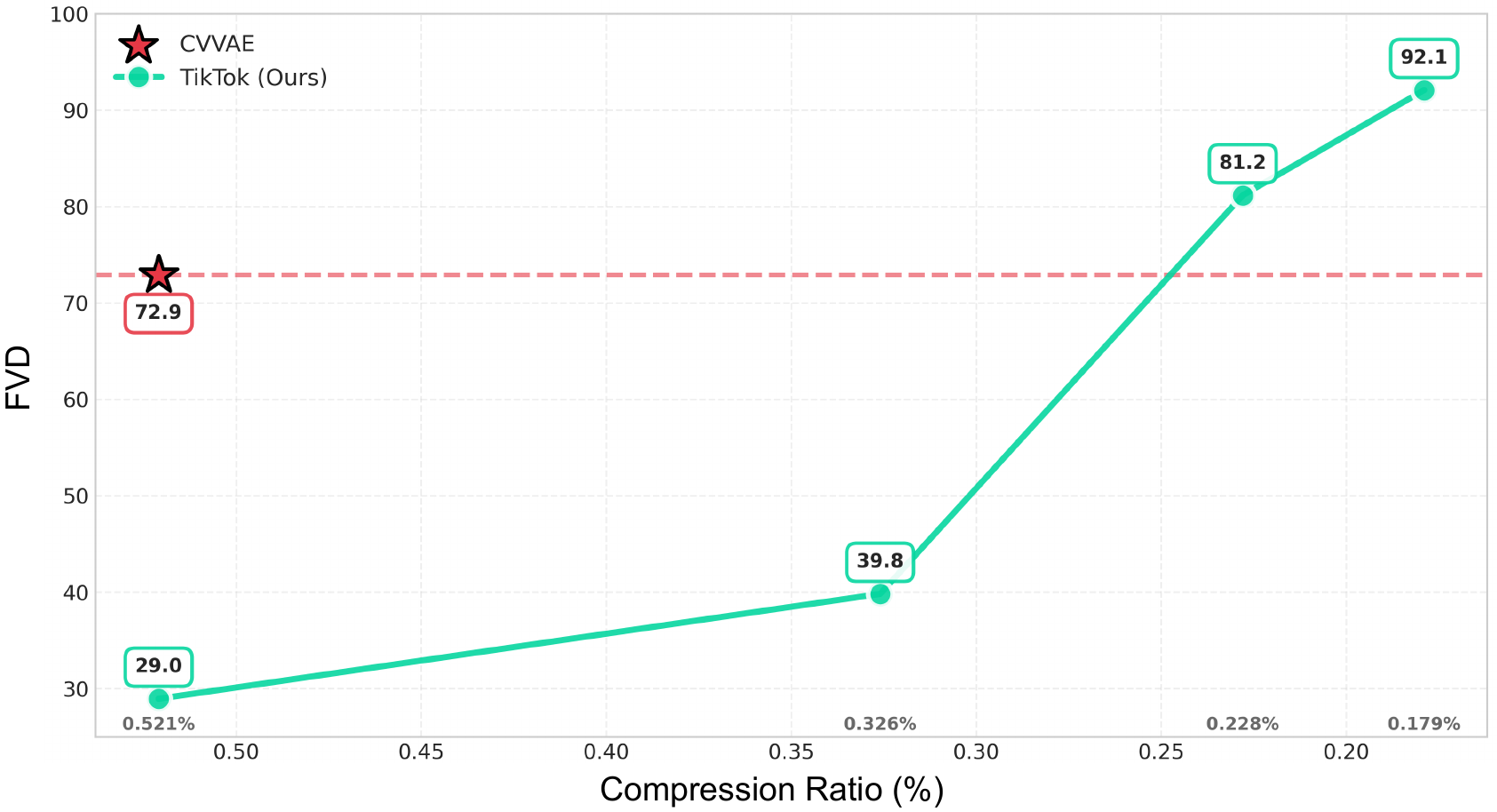}
    \caption{Trend induced by TIV-token capacity. The visualization complements Table~\ref{tab:multipletiv} by showing the quality--efficiency trade-off as the number of TIV tokens changes.}
    \label{fig:multipletiv}
\end{figure}

\begin{table*}[t]
\centering
\caption{Quantitative effect of TIV/TV token allocation. Experiments are conducted at $128\times256\times256$ resolution; the table reports exact reconstruction and compression metrics.}
\label{tab:TIV_TV_ratio}
\setlength{\tabcolsep}{5pt}
\begin{tabular*}{\textwidth}{@{\extracolsep{\fill}}lcccccc}
\toprule
Method & TIV:TV Ratio & Comp. Rate (\%)$\downarrow$ & PSNR $\uparrow$ & SSIM $\uparrow$ & LPIPS $\downarrow$ & rFVD $\downarrow$ \\
\midrule
CV-VAE & - & 0.521 & 29.00 & 0.8831 & 0.0729 & 72.91 \\
TivTok & 1:3 & 0.318 & 28.24 & 0.8663 & 0.0761 & 41.33 \\
TivTok & 1:1 & 0.229 & 27.52 & 0.8503 & 0.0887 & 64.76 \\
TivTok & 3:1 & 0.179 & 26.23 & 0.8210 & 0.1057 & 92.09 \\
\bottomrule
\end{tabular*}
\end{table*}

\begin{figure}[t]
    \centering
    \includegraphics[width=\linewidth]{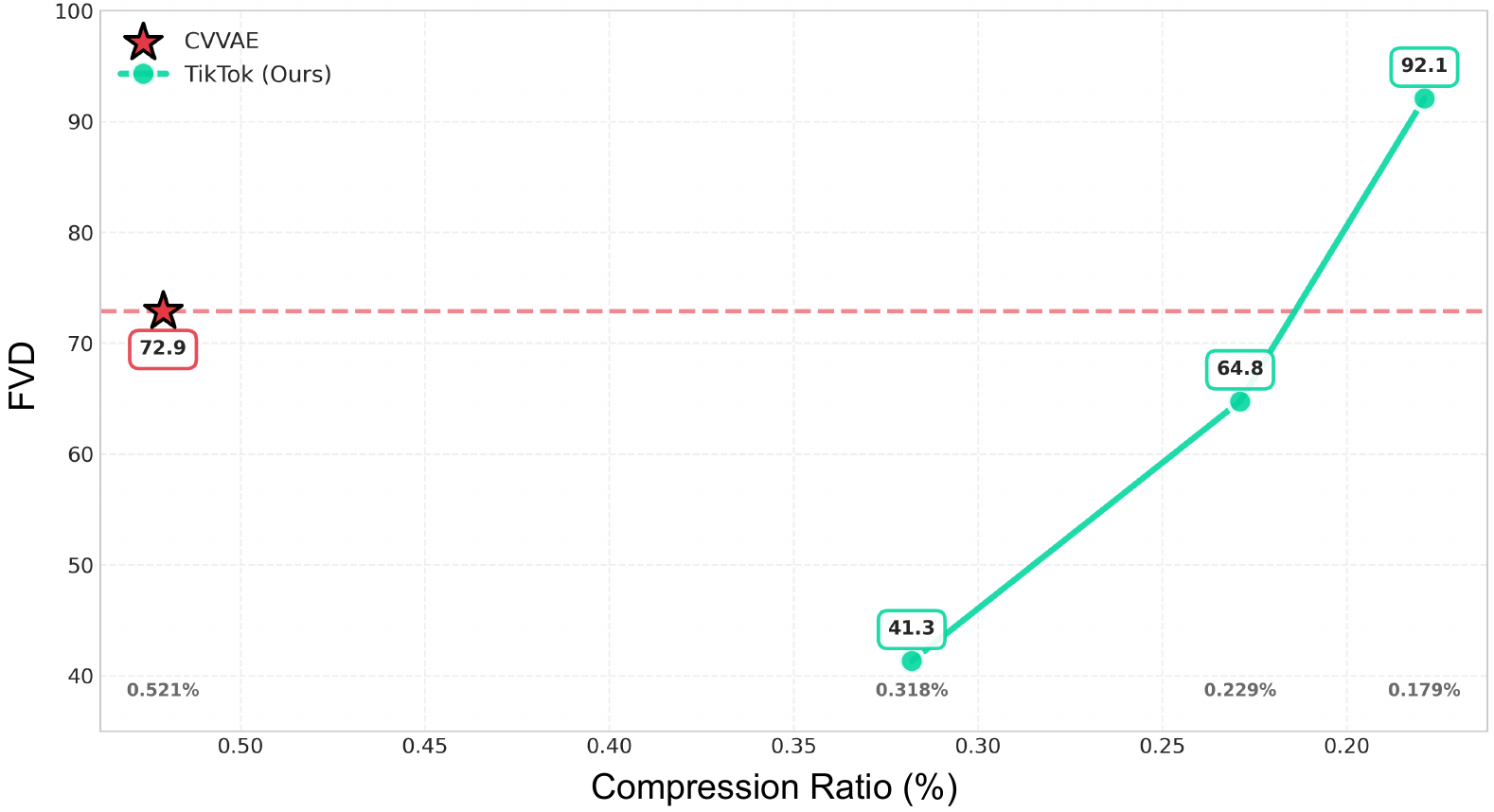}
    \caption{Trend induced by TIV/TV allocation. The visualization complements Table~\ref{tab:TIV_TV_ratio} by showing how token allocation changes the balance between reconstruction quality and compression efficiency.}
    \label{fig:tivratio}
\end{figure}

We next study how much capacity should be assigned to the invariant branch. Tables~\ref{tab:multipletiv} and~\ref{tab:TIV_TV_ratio} report the exact reconstruction and compression metrics, while Figures~\ref{fig:multipletiv} and~\ref{fig:tivratio} visualize the corresponding trade-offs. This separation avoids relying on a single scalar score: the tables provide precise operating points, and the figures make the quality--efficiency trend easier to inspect.

The first study varies the number of TIV tokens. As shown in Table~\ref{tab:multipletiv} and Figure~\ref{fig:multipletiv}, increasing TIV capacity improves reconstruction quality because the model can store richer reusable structure, but it also increases the token budget and weakens compression efficiency. The second study varies the TIV/TV allocation. Table~\ref{tab:TIV_TV_ratio} and Figure~\ref{fig:tivratio} show that assigning more capacity to TV tokens improves frame-specific detail modeling, whereas assigning more capacity to TIV tokens favors compact long-video representation. These results suggest that TivTok should allocate enough invariant capacity to capture reusable semantics, while leaving sufficient TV capacity for fast-changing residuals.

\subsection{Ablation Study}
\begin{table*}[t]
\centering
\caption{Ablation studies on the proposed 
components of TivTok.}
\label{tab:ablation}
\setlength{\tabcolsep}{5pt}
\begin{tabular*}{\textwidth}{@{\extracolsep{\fill}}lcccc}
    \toprule
    Methods & PSNR$\uparrow$ & SSIM$\uparrow$ & 
    LPIPS$\downarrow$ & rFVD$\downarrow$ \\
    \midrule
    w/o TIV/TV factorization & 27.24 & 0.8530 & 
    0.0748 & 91.99 \\
    w/o Scope-Induced Factorization (SIF) & 19.67 & 
    0.5691 & 0.5691 & 1359.38 \\
    w/o Invariant Broadcasting (IB) & 17.69 & 0.4665 
    & 0.6083 & 3694.34 \\
    w/o Cross-Chunk TIV Reuse & 25.81 & 0.8219 & 
    0.1069 & 93.49 \\
    \midrule
    TivTok & \textbf{29.05} & \textbf{0.8831} & 
    \textbf{0.0719} & \textbf{38.49} \\
    \bottomrule
\end{tabular*}
\end{table*}

Table~\ref{tab:ablation} presents ablation studies on the $32{\times}256{\times}256$ setting. Removing TIV/TV factorization leads to significant performance degradation, confirming that holistic tokenization struggles to exploit the persistent structure inherent to videos. Removing Scope-Induced Factorization (SIF) or Invariant Broadcasting (IB) causes severe reconstruction collapse, highlighting that these two components are mutually dependent: SIF structures the encoder to produce factorized representations, while IB relies on this structure for parallel decoding. Ablating Cross-Chunk TIV Reuse results in noticeable but recoverable performance drops, suggesting that temporal invariants are indeed shared across chunks and that explicit cross-chunk reuse further enhances their extraction and utilization. Together, these results validate that each proposed component addresses a distinct and necessary aspect of scalable video tokenization.

\subsection{Scalability of TivTok}
\begin{table*}[t]
    \caption{Scalability of TivTok. TivTok consistently improves with larger models and datasets, and maintains strong performance across different video resolutions.}
    \label{tab:scalability}
    \centering
    \setlength{\tabcolsep}{4pt}
    \begin{tabular*}{\textwidth}{@{\extracolsep{\fill}}lccccc}
        \toprule
        \textbf{Method} & \textbf{Comp. Rate (\%)}$\downarrow$ &\textbf{PSNR}$\uparrow$ & \textbf{SSIM}$\uparrow$ & \textbf{LPIPS}$\downarrow$  & \textbf{rFVD}$\downarrow$ \\
        \midrule
        \multicolumn{6}{l}{\textit{Scalability of Model Size. Tested on UCF-101}} \\
        \hdashline 
        \addlinespace
        TivTok-Small  &0.52 & 27.73 & 0.861 &0.081 &47.31\\
        TivTok-Base  &0.52 & 30.13 & 0.901 & 0.061 &21.29 \\
        TivTok-Large  &0.52 &30.94 &0.912 &0.049 &13.11 \\
        \midrule
        \multicolumn{6}{l}{\textit{Scalability of Dataset Size and Resolution}} \\
        \hdashline
        \addlinespace
        CMD-WebVid-256~\citep{yu2024efficientcmd}   &6.85& 26.55 &0.795 &0.110 & 98.62\\
        HiVAE-WebVid-256~\citep{liu2025hivae}  &0.27 & 29.35 &0.834 & 0.096 &61.94\\
        TivTok-WebVid-256 &0.26 & 28.61 & 0.829 & 0.073 &22.96\\
        TivTok-WebVid-256 &0.52 & 31.69 & 0.896 & 0.048 & 7.15 \\
        TivTok-VidProM-256  &0.52 & 33.17 & 0.938 & 0.028 &5.63\\
        TivTok-VidProM-512 &0.52 &33.56 &0.937 &0.043 &9.08 \\
        \bottomrule
    \end{tabular*}
\end{table*}

Table~\ref{tab:scalability} evaluates the scalability of our method with respect to model size, dataset scale, and resolution. Performance improves consistently as model size increases, indicating that larger models more effectively capture complex spatiotemporal dynamics. Expanding the training data—e.g., with WebVid-10M~\citep{bain2021webvid10m} and VidProM~\citep{wang2024vidprom}—further enhances performance by providing greater diversity and coverage. Compared with other decomposition-based video tokenizers such as CMD~\citep{yu2024efficientcmd} and HiVAE~\citep{liu2025hivae}, our method avoids manually imposing fixed decomposition patterns. Instead, it explicitly extracts and reuses time-invariant information, leading to superior performance on WebVid-10M~\citep{bain2021webvid10m} and demonstrating stronger scalability. Regarding resolution, experiments on VidProM show that our approach remains effective for higher-resolution video data, highlighting its robustness and potential for large-scale, high-resolution scenarios. Overall, these results confirm that our method scales effectively across model size, dataset scale, and resolution, enabled by the general principle of reusing time-invariant information and a clean, streamlined design.

\section{Discussion}

TivTok is most beneficial when videos contain reusable structure over many frames. SIF encourages TIV tokens to aggregate shared semantics, while IB reuses these tokens across frames and chunks. As video length increases, this reuse avoids repeatedly encoding the same persistent content, explaining the larger efficiency gain observed in the long-video setting. The token visualizations further show that the reusable component is not limited to pixel-static background, but can include semantic structure that remains stable under motion.

The main limitation is that the assumption of reusable temporal invariants may be weaker for videos with abrupt scene cuts, highly non-stationary camera motion, rapidly changing objects, or little persistent content. In such cases, fewer tokens can be safely shared across time and the benefit of broadcasting may decrease. Jointly optimizing TivTok with larger downstream video generation models is also left for future work.

\section{Conclusion}
We present TivTok (\emph{Time-Invariant Tokenizer}), a reuse-aware video tokenizer that represents persistent information with Time-Invariant (TIV) tokens and frame-specific residuals with Time-Variant (TV) tokens. TivTok realizes this representation through two complementary mechanisms. Scope-Induced Factorization (SIF) assigns different attention scopes to the two token groups, encouraging TIV tokens to aggregate information shared across frames while keeping TV tokens focused on frame-local variation. Invariant Broadcasting (IB) reuses the same TIV tokens during reconstruction and across chunks, enabling parallel decoding and long-video tokenization with a smaller token budget. Our analysis and visualizations show that TIV tokens capture reusable semantic structure beyond pixel-level persistence, supporting their reuse across frames and chunks. Experiments show that TivTok achieves an rFVD of 12.65 on the standard $16{\times}256{\times}256$ benchmark and improves compression efficiency by $2.91{\times}$ for 128-frame videos compared with the evaluated baselines. These results suggest that separating reusable and frame-specific content provides a practical direction for scalable video tokenization.

{\footnotesize
  \bibliographystyle{spbasic}
  \bibliography{main}

\begin{thebibliography}{58}
\providecommand{\natexlab}[1]{#1}
\providecommand{\url}[1]{{#1}}
\providecommand{\urlprefix}{URL }
\expandafter\ifx\csname urlstyle\endcsname\relax
  \providecommand{\doi}[1]{DOI~\discretionary{}{}{}#1}\else
  \providecommand{\doi}{DOI~\discretionary{}{}{}\begingroup
  \urlstyle{rm}\Url}\fi
\providecommand{\eprint}[2][]{\url{#2}}

\bibitem[{Agarwal et~al.(2025)Agarwal, Ali, Bala, Balaji, Barker, Cai,
  Chattopadhyay, Chen, Cui, Ding et~al.}]{agarwal2025cosmos}
Agarwal N, Ali A, Bala M, Balaji Y, Barker E, Cai T, Chattopadhyay P, Chen Y,
  Cui Y, Ding Y, et~al. (2025) Cosmos world foundation model platform for
  physical ai. arXiv preprint arXiv:250103575

\bibitem[{Bain et~al.(2021)Bain, Nagrani, Varol, and
  Zisserman}]{bain2021webvid10m}
Bain M, Nagrani A, Varol G, Zisserman A (2021) Frozen in time: A joint video
  and image encoder for end-to-end retrieval. In: Proceedings of the IEEE/CVF
  international conference on computer vision, pp 1728--1738

\bibitem[{Blattmann et~al.(2023{\natexlab{a}})Blattmann, Dockhorn, Kulal,
  Mendelevitch, Kilian, Lorenz, Levi, English, Voleti, Letts
  et~al.}]{blattmann2023stablevideodiffusion}
Blattmann A, Dockhorn T, Kulal S, Mendelevitch D, Kilian M, Lorenz D, Levi Y,
  English Z, Voleti V, Letts A, et~al. (2023{\natexlab{a}}) Stable video
  diffusion: Scaling latent video diffusion models to large datasets. arXiv
  preprint arXiv:231115127

\bibitem[{Blattmann et~al.(2023{\natexlab{b}})Blattmann, Rombach, Ling,
  Dockhorn, Kim, Fidler, and Kreis}]{blattmann2023alignyourlatents}
Blattmann A, Rombach R, Ling H, Dockhorn T, Kim SW, Fidler S, Kreis K
  (2023{\natexlab{b}}) Align your latents: High-resolution video synthesis with
  latent diffusion models. In: Proceedings of the IEEE/CVF conference on
  computer vision and pattern recognition, pp 22563--22575

\bibitem[{Carreira et~al.(2018)Carreira, Noland, Banki-Horvath, Hillier, and
  Zisserman}]{carreira2018shortk600}
Carreira J, Noland E, Banki-Horvath A, Hillier C, Zisserman A (2018) A short
  note about kinetics-600. arXiv preprint arXiv:180801340

\bibitem[{Chen et~al.(2025{\natexlab{a}})Chen, Wang, Li, Sun, Chen, Liu, Wang,
  Raj, Liu, and Barsoum}]{chen2025softvq}
Chen H, Wang Z, Li X, Sun X, Chen F, Liu J, Wang J, Raj B, Liu Z, Barsoum E
  (2025{\natexlab{a}}) Softvq-vae: Efficient 1-dimensional continuous
  tokenizer. In: Proceedings of the Computer Vision and Pattern Recognition
  Conference, pp 28358--28370

\bibitem[{Chen et~al.(2024{\natexlab{a}})Chen, Li, Lin, Zhu, Wang, Yuan, Zhou,
  Cheng, and Yuan}]{chen2024odvae}
Chen L, Li Z, Lin B, Zhu B, Wang Q, Yuan S, Zhou X, Cheng X, Yuan L
  (2024{\natexlab{a}}) Od-vae: An omni-dimensional video compressor for
  improving latent video diffusion model. arXiv preprint arXiv:240901199

\bibitem[{Chen et~al.(2024{\natexlab{b}})Chen, Liu, Wu, Sun, Lu, and
  Duan}]{chen2024dreamcinema}
Chen W, Liu F, Wu D, Sun H, Lu J, Duan Y (2024{\natexlab{b}}) Dreamcinema:
  Cinematic transfer with free camera and 3d character. arXiv preprint
  arXiv:240812601

\bibitem[{Chen et~al.(2025{\natexlab{b}})Chen, Bi, Huang, Zheng, and
  Duan}]{chen2025scenecompleter}
Chen W, Bi J, Huang Y, Zheng W, Duan Y (2025{\natexlab{b}}) Scenecompleter:
  Dense 3d scene completion for generative novel view synthesis. arXiv preprint
  arXiv:250610981

\bibitem[{De~Rivaz and Haughton(2019)}]{de2019av1}
De~Rivaz P, Haughton J (2019) Av1 bitstream \& decoding process specification.
  The Alliance for Open Media 1:2

\bibitem[{Goodfellow et~al.(2020)Goodfellow, Pouget-Abadie, Mirza, Xu,
  Warde-Farley, Ozair, Courville, and
  Bengio}]{goodfellow2020generativeadversualloss}
Goodfellow I, Pouget-Abadie J, Mirza M, Xu B, Warde-Farley D, Ozair S,
  Courville A, Bengio Y (2020) Generative adversarial networks. Communications
  of the ACM 63(11):139--144

\bibitem[{HaCohen et~al.(2024)HaCohen, Chiprut, Brazowski, Shalem, Moshe,
  Richardson, Levin, Shiran, Zabari, Gordon et~al.}]{hacohen2024ltx}
HaCohen Y, Chiprut N, Brazowski B, Shalem D, Moshe D, Richardson E, Levin E,
  Shiran G, Zabari N, Gordon O, et~al. (2024) Ltx-video: Realtime video latent
  diffusion. arXiv preprint arXiv:250100103

\bibitem[{Huang et~al.(2025{\natexlab{a}})Huang, Huang, Wang, Lin, Ning, Wan,
  Zhang, Wang, and Liu}]{huang2025filmaster}
Huang K, Huang Y, Wang X, Lin Z, Ning X, Wan P, Zhang D, Wang Y, Liu X
  (2025{\natexlab{a}}) Filmaster: Bridging cinematic principles and generative
  ai for automated film generation. arXiv preprint arXiv:250618899

\bibitem[{Huang et~al.(2024)Huang, Zheng, Gao, Tao, Wan, Zhang, Zhou, and
  Lu}]{huang2024owl}
Huang Y, Zheng W, Gao Y, Tao X, Wan P, Zhang D, Zhou J, Lu J (2024) Owl-1: Omni
  world model for consistent long video generation. arXiv preprint
  arXiv:241209600

\bibitem[{Huang et~al.(2025{\natexlab{b}})Huang, Chen, Zheng, Duan, Zhou, and
  Lu}]{huang2025spectralar}
Huang Y, Chen W, Zheng W, Duan Y, Zhou J, Lu J (2025{\natexlab{b}}) Spectralar:
  Spectral autoregressive visual generation. arXiv preprint arXiv:250610962

\bibitem[{Jang et~al.(2025)Jang, Yu, Shin, Abbeel, and
  Seo}]{jang2025efficientcoordtok}
Jang H, Yu S, Shin J, Abbeel P, Seo Y (2025) Efficient long video tokenization
  via coordinate-based patch reconstruction. In: Proceedings of the Computer
  Vision and Pattern Recognition Conference, pp 22853--22863

\bibitem[{Jin et~al.(2024)Jin, Sun, Xu, Chen, Jiang, Huang, Song, Liu, Zhang,
  Song et~al.}]{jin2024videolavit}
Jin Y, Sun Z, Xu K, Chen L, Jiang H, Huang Q, Song C, Liu Y, Zhang D, Song Y,
  et~al. (2024) Video-lavit: Unified video-language pre-training with decoupled
  visual-motional tokenization. arXiv preprint arXiv:240203161

\bibitem[{Johnson et~al.(2016)Johnson, Alahi, and
  Fei-Fei}]{johnson2016perceptualloss}
Johnson J, Alahi A, Fei-Fei L (2016) Perceptual losses for real-time style
  transfer and super-resolution. In: European conference on computer vision,
  Springer, pp 694--711

\bibitem[{Kim et~al.(2024)Kim, Lee, Park, Kim, Lee, Kim, and
  Yoo}]{kim2024hybridhvdm}
Kim K, Lee H, Park J, Kim S, Lee K, Kim S, Yoo J (2024) Hybrid video diffusion
  models with 2d triplane and 3d wavelet representation. In: European
  Conference on Computer Vision, Springer, pp 148--165

\bibitem[{Larsen et~al.(2016)Larsen, S{\o}nderby, Larochelle, and
  Winther}]{larsen2016autoencodingperceptualloss}
Larsen ABL, S{\o}nderby SK, Larochelle H, Winther O (2016) Autoencoding beyond
  pixels using a learned similarity metric. In: International conference on
  machine learning, PMLR, pp 1558--1566

\bibitem[{Li et~al.(2025)Li, Tian, Xia, Liao, Guo, Yan, Li, Dai, Li, and
  Yang}]{li2025learningadaptok}
Li Y, Tian C, Xia R, Liao N, Guo W, Yan J, Li H, Dai J, Li H, Yang X (2025)
  Learning adaptive and temporally causal video tokenization in a 1d latent
  space. arXiv preprint arXiv:250517011

\bibitem[{Li et~al.(2024)Li, Zhang, Lin, Xiong, Long, Deng, Zhang, Liu, Huang,
  Xiao et~al.}]{li2024hunyuan}
Li Z, Zhang J, Lin Q, Xiong J, Long Y, Deng X, Zhang Y, Liu X, Huang M, Xiao Z,
  et~al. (2024) Hunyuan-dit: A powerful multi-resolution diffusion transformer
  with fine-grained chinese understanding. arXiv preprint arXiv:240508748

\bibitem[{Liu et~al.(2024)Liu, Wang, Chen, Sun, and Duan}]{liu2024makeyour3d}
Liu F, Wang H, Chen W, Sun H, Duan Y (2024) Make-your-3d: Fast and consistent
  subject-driven 3d content generation. In: European Conference on Computer
  Vision, Springer, pp 389--406

\bibitem[{Liu et~al.(2025)Liu, Sun, Zhang, Di, Gong, Li, Wei, and
  Zou}]{liu2025hivae}
Liu H, Sun W, Zhang Q, Di D, Gong B, Li H, Wei C, Zou C (2025) Hi-vae:
  Efficient video autoencoding with global and detailed motion. arXiv preprint
  arXiv:250607136

\bibitem[{Ma et~al.(2024)Ma, Goldstein, Albergo, Boffi, Vanden-Eijnden, and
  Xie}]{ma2024sit}
Ma N, Goldstein M, Albergo MS, Boffi NM, Vanden-Eijnden E, Xie S (2024) Sit:
  Exploring flow and diffusion-based generative models with scalable
  interpolant transformers. In: European Conference on Computer Vision,
  Springer, pp 23--40

\bibitem[{Podell et~al.(2023)Podell, English, Lacey, Blattmann, Dockhorn,
  M{\"u}ller, Penna, and Rombach}]{podell2023sdxl}
Podell D, English Z, Lacey K, Blattmann A, Dockhorn T, M{\"u}ller J, Penna J,
  Rombach R (2023) Sdxl: Improving latent diffusion models for high-resolution
  image synthesis. arXiv preprint arXiv:230701952

\bibitem[{Ren et~al.(2024{\natexlab{a}})Ren, Huang, Zeng, Museth, Fidler, and
  Williams}]{ren2024xcube}
Ren X, Huang J, Zeng X, Museth K, Fidler S, Williams F (2024{\natexlab{a}})
  Xcube: Large-scale 3d generative modeling using sparse voxel hierarchies. In:
  Proceedings of the IEEE/CVF conference on computer vision and pattern
  recognition, pp 4209--4219

\bibitem[{Ren et~al.(2024{\natexlab{b}})Ren, Lu, Liang, Wu, Ling, Chen, Fidler,
  Williams, and Huang}]{ren2024scube}
Ren X, Lu Y, Liang H, Wu Z, Ling H, Chen M, Fidler S, Williams F, Huang J
  (2024{\natexlab{b}}) Scube: Instant large-scale scene reconstruction using
  voxsplats. Advances in Neural Information Processing Systems 37:97670--97698

\bibitem[{Richardson(2004)}]{richardson2004hh264mpeg4}
Richardson IE (2004) H. 264 and MPEG-4 video compression: video coding for
  next-generation multimedia. John Wiley \& Sons

\bibitem[{Rombach et~al.(2022)Rombach, Blattmann, Lorenz, Esser, and
  Ommer}]{rombach2022highlatentdiffusion}
Rombach R, Blattmann A, Lorenz D, Esser P, Ommer B (2022) High-resolution image
  synthesis with latent diffusion models. In: Proceedings of the IEEE/CVF
  conference on computer vision and pattern recognition, pp 10684--10695

\bibitem[{Shu et~al.(2026)Shu, Qiu, Yao, and Mei}]{shu2026guidedvdm}
Shu Y, Qiu Z, Yao T, Mei T (2026) Guidedvdm: Controllable video generation with
  long-term consistency. International Journal of Computer Vision 134(6),
  \doi{10.1007/s11263-026-02901-4}

\bibitem[{Soomro et~al.(2012)Soomro, Zamir, and Shah}]{soomro2012ucf101}
Soomro K, Zamir AR, Shah M (2012) Ucf101: A dataset of 101 human actions
  classes from videos in the wild. arXiv preprint arXiv:12120402

\bibitem[{Tan et~al.(2024)Tan, Xue, Jia, Wang, Ye, Shi, Sun, Wu, Chen, and
  Jiang}]{tan2024sweettok}
Tan Z, Xue B, Jia J, Wang J, Ye W, Shi S, Sun M, Wu W, Chen Q, Jiang P (2024)
  Sweettok: Semantic-aware spatial-temporal tokenizer for compact video
  discretization. arXiv preprint arXiv:241210443

\bibitem[{Tang et~al.(2024)Tang, He, Guo, Cheng, Song, and
  Bian}]{tang2024vidtok}
Tang A, He T, Guo J, Cheng X, Song L, Bian J (2024) Vidtok: A versatile and
  open-source video tokenizer. arXiv preprint arXiv:241213061

\bibitem[{Tian et~al.(2024{\natexlab{a}})Tian, Jiang, Yuan, Peng, and
  Wang}]{tian2024visualvar}
Tian K, Jiang Y, Yuan Z, Peng B, Wang L (2024{\natexlab{a}}) Visual
  autoregressive modeling: Scalable image generation via next-scale prediction.
  Advances in neural information processing systems 37:84839--84865

\bibitem[{Tian et~al.(2024{\natexlab{b}})Tian, Dai, Bao, Qiu, Yang, Luo, Wu,
  and Jiang}]{tian2024reducio}
Tian R, Dai Q, Bao J, Qiu K, Yang Y, Luo C, Wu Z, Jiang YG (2024{\natexlab{b}})
  Reducio! generating 1k video within 16 seconds using extremely compressed
  motion latents. arXiv preprint arXiv:241113552

\bibitem[{Unterthiner et~al.(2018)Unterthiner, Van~Steenkiste, Kurach,
  Marinier, Michalski, and Gelly}]{unterthiner2018towardsrfvd}
Unterthiner T, Van~Steenkiste S, Kurach K, Marinier R, Michalski M, Gelly S
  (2018) Towards accurate generative models of video: A new metric \&
  challenges. arXiv preprint arXiv:181201717

\bibitem[{Wang et~al.(2024{\natexlab{a}})Wang, Suri, Ren, Chen, and
  Shrivastava}]{wang2024larp}
Wang H, Suri S, Ren Y, Chen H, Shrivastava A (2024{\natexlab{a}}) Larp:
  Tokenizing videos with a learned autoregressive generative prior. arXiv
  preprint arXiv:241021264

\bibitem[{Wang et~al.(2024{\natexlab{b}})Wang, Jiang, Yuan, Peng, Wu, and
  Jiang}]{wang2024omnitokenizer}
Wang J, Jiang Y, Yuan Z, Peng B, Wu Z, Jiang YG (2024{\natexlab{b}})
  Omnitokenizer: A joint image-video tokenizer for visual generation. Advances
  in Neural Information Processing Systems 37:28281--28295

\bibitem[{Wang et~al.(2026)Wang, Shen, Xiao, Tian, Wang, Hu, Zhu, and
  Feng}]{wang2026breakingredundancy}
Wang S, Shen L, Xiao J, Tian Z, Wang F, Hu X, Zhu Y, Feng G (2026) Breaking
  redundancy via 3d sparse geometry: 3d-aware neural compression for multi-view
  videos. International Journal of Computer Vision 134(1),
  \doi{10.1007/s11263-025-02604-2}

\bibitem[{Wang and Yang(2024)}]{wang2024vidprom}
Wang W, Yang Y (2024) Vidprom: A million-scale real prompt-gallery dataset for
  text-to-video diffusion models. Advances in Neural Information Processing
  Systems 37:65618--65642

\bibitem[{Wang et~al.(2025)Wang, Guo, Xie, He, Sun, and Bian}]{wang2025vidtwin}
Wang Y, Guo J, Xie X, He T, Sun X, Bian J (2025) Vidtwin: Video vae with
  decoupled structure and dynamics. In: Proceedings of the Computer Vision and
  Pattern Recognition Conference, pp 22922--22932

\bibitem[{Wang et~al.(2004)Wang, Bovik, Sheikh, and Simoncelli}]{wang2004ssim}
Wang Z, Bovik AC, Sheikh HR, Simoncelli EP (2004) Image quality assessment:
  from error visibility to structural similarity. IEEE transactions on image
  processing 13(4):600--612

\bibitem[{Wu et~al.(2024)Wu, Yin, Feng, He, Li, Hao, and
  Long}]{wu2024ivideogpt}
Wu J, Yin S, Feng N, He X, Li D, Hao J, Long M (2024) ivideogpt: Interactive
  videogpts are scalable world models. Advances in Neural Information
  Processing Systems 37:68082--68119

\bibitem[{Xu et~al.(2023)Xu, Wang, Wang, Yu, and
  Jia}]{xu2023conditionaltemporalvae}
Xu X, Wang Y, Wang L, Yu B, Jia J (2023) Conditional temporal variational
  autoencoder for action video prediction. International Journal of Computer
  Vision 131(10):2699--2722, \doi{10.1007/s11263-023-01832-8}

\bibitem[{Yan et~al.(2024)Yan, Mnih, Faust, Zaharia, Abbeel, and
  Liu}]{yan2024elastictok}
Yan W, Mnih V, Faust A, Zaharia M, Abbeel P, Liu H (2024) Elastictok: Adaptive
  tokenization for image and video. arXiv preprint arXiv:241008368

\bibitem[{Yang et~al.(2021)Yang, Shen, and Zhou}]{yang2021semantichierarchy}
Yang C, Shen Y, Zhou B (2021) Semantic hierarchy emerges in deep generative
  representations for scene synthesis. International Journal of Computer Vision
  129(5):1451--1466, \doi{10.1007/s11263-020-01429-5}

\bibitem[{Yao et~al.(2025)Yao, Yang, and Wang}]{yao2025lightningdit}
Yao J, Yang B, Wang X (2025) Reconstruction vs. generation: Taming optimization
  dilemma in latent diffusion models. In: Proceedings of the Computer Vision
  and Pattern Recognition Conference, pp 15703--15712

\bibitem[{Yoo et~al.(2023)Yoo, Kim, Lee, Kim, and Hong}]{yoo2023towardsMeBT}
Yoo J, Kim S, Lee D, Kim C, Hong S (2023) Towards end-to-end generative
  modeling of long videos with memory-efficient bidirectional transformers. In:
  Proceedings of the IEEE/CVF Conference on Computer Vision and Pattern
  Recognition, pp 22888--22897

\bibitem[{Yu et~al.(2024{\natexlab{a}})Yu, Weber, Deng, Shen, Cremers, and
  Chen}]{yu2024imagetitok}
Yu Q, Weber M, Deng X, Shen X, Cremers D, Chen LC (2024{\natexlab{a}}) An image
  is worth 32 tokens for reconstruction and generation. Advances in Neural
  Information Processing Systems 37:128940--128966

\bibitem[{Yu et~al.(2023)Yu, Sohn, Kim, and Shin}]{yu2023videopvdm}
Yu S, Sohn K, Kim S, Shin J (2023) Video probabilistic diffusion models in
  projected latent space. In: Proceedings of the IEEE/CVF conference on
  computer vision and pattern recognition, pp 18456--18466

\bibitem[{Yu et~al.(2024{\natexlab{b}})Yu, Nie, Huang, Li, Shin, and
  Anandkumar}]{yu2024efficientcmd}
Yu S, Nie W, Huang DA, Li B, Shin J, Anandkumar A (2024{\natexlab{b}})
  Efficient video diffusion models via content-frame motion-latent
  decomposition. arXiv preprint arXiv:240314148

\bibitem[{Yu et~al.(2025)Yu, Hahn, Kondratyuk, Shin, Gupta, Lezama, Essa, Ross,
  and Huang}]{yu2025maltdiffusion}
Yu S, Hahn M, Kondratyuk D, Shin J, Gupta A, Lezama J, Essa I, Ross D, Huang J
  (2025) Malt diffusion: Memory-augmented latent transformers for any-length
  video generation. arXiv preprint arXiv:250212632

\bibitem[{Zhang et~al.(2018)Zhang, Isola, Efros, Shechtman, and
  Wang}]{zhang2018unreasonablelpips}
Zhang R, Isola P, Efros AA, Shechtman E, Wang O (2018) The unreasonable
  effectiveness of deep features as a perceptual metric. In: Proceedings of the
  IEEE conference on computer vision and pattern recognition, pp 586--595

\bibitem[{Zhang et~al.(2025)Zhang, Wang, Chen, Qin, Hao, Mei, and
  Zhu}]{zhang2025scenariodiff}
Zhang Y, Wang X, Chen H, Qin C, Hao Y, Mei H, Zhu W (2025) Scenariodiff:
  Text-to-video generation with dynamic transformations of scene conditions.
  International Journal of Computer Vision 133(7):4909--4922,
  \doi{10.1007/s11263-025-02413-7}

\bibitem[{Zhao et~al.(2024)Zhao, Zhang, Cun, Yang, Niu, Li, Hu, and
  Shan}]{zhao2024cvvae}
Zhao S, Zhang Y, Cun X, Yang S, Niu M, Li X, Hu W, Shan Y (2024) Cv-vae: A
  compatible video vae for latent generative video models. Advances in Neural
  Information Processing Systems 37:12847--12871

\bibitem[{Zheng et~al.(2024{\natexlab{a}})Zheng, Chen, Huang, Zhang, Duan, and
  Lu}]{zheng2024occworld}
Zheng W, Chen W, Huang Y, Zhang B, Duan Y, Lu J (2024{\natexlab{a}}) Occworld:
  Learning a 3d occupancy world model for autonomous driving. In: European
  conference on computer vision, Springer, pp 55--72

\bibitem[{Zheng et~al.(2024{\natexlab{b}})Zheng, Peng, Yang, Shen, Li, Liu,
  Zhou, Li, and You}]{zheng2024opensora}
Zheng Z, Peng X, Yang T, Shen C, Li S, Liu H, Zhou Y, Li T, You Y
  (2024{\natexlab{b}}) Open-sora: Democratizing efficient video production for
  all. arXiv preprint arXiv:241220404

\end{thebibliography}
}

\end{document}